\documentclass[sigconf]{acmart}

\AtBeginDocument{%
  \providecommand\BibTeX{{%
    \normalfont B\kern-0.5em{\scshape i\kern-0.25em b}\kern-0.8em\TeX}}}

\copyrightyear{2022}
\acmYear{2022}
\setcopyright{acmcopyright}\acmConference[MM '22]{Proceedings of the 30th ACM
International Conference on Multimedia}{October 10--14, 2022}{Lisboa, Portugal}
\acmBooktitle{Proceedings of the 30th ACM International Conference on Multimedia
(MM '22), October 10--14, 2022, Lisboa, Portugal}
\acmPrice{15.00}
\acmDOI{10.1145/3503161.3547995}
\acmISBN{978-1-4503-9203-7/22/10}

\acmSubmissionID{1014}
\usepackage{xcolor}
\usepackage{bbding} 
\usepackage{algorithm}

\newcommand{\mypm}{$\pm$\,}

\begin{document}

\title{ME-D2N: Multi-Expert Domain Decompositional Network for Cross-Domain Few-Shot Learning}

\author{Yuqian Fu$^{\ast}$ \country{}}
\affiliation{fuyq20@fudan.edu.cn \country{}}
\affiliation{
Shanghai Key Lab of Intelligent Information Processing, School of Computer Science, Fudan University 
\country{}
}

\author{Yu Xie$^{\ast}$ \country{}}
\author{Yanwei Fu \country{}}
 \affiliation{\{yxie18, yanweifu\}@fudan.edu.cn \country{}}
\affiliation{School of Data Science, Fudan University
\country{}
}

\author{Jingjing Chen\country{}}
\author{Yu-Gang Jiang\#\country{}}
 \affiliation{\{chenjingjing, ygj\}@fudan.edu.cn \country{}}
\affiliation{Shanghai Key Lab of Intelligent Information Processing, School of Computer Science, Fudan University 
\country{}
}

\thanks{$\ast$ indicates equal contributions, $\#$ indicates corresponding author}

\renewcommand{\shortauthors}{Yuqian Fu et al.}

\begin{abstract}
Recently, Cross-Domain Few-Shot Learning (CD-FSL) which aims at addressing the Few-Shot Learning (FSL) problem across different domains has attracted rising attention. The core challenge of CD-FSL lies in the domain gap between the source and novel target datasets. Though many attempts have been made for CD-FSL without any target data during model training, the huge domain gap makes it still hard for existing CD-FSL methods to achieve very satisfactory results.
Alternatively, learning CD-FSL models with few labeled target domain data which is more realistic and promising is advocated in previous work~\cite{fu2021meta}. Thus, in this paper, we stick to this setting and technically contribute a novel Multi-Expert Domain Decompositional Network (ME-D2N).
Concretely, to solve the data imbalance problem between the source data with sufficient examples and the auxiliary target data with limited examples, we build our model under the umbrella of multi-expert learning. 
Two teacher models which can be considered to be experts in their corresponding domain are first trained on the source and the auxiliary target sets, respectively. 
Then, the knowledge distillation technique is introduced to transfer the knowledge from two teachers to a unified student model.
Taking a step further, to help our student model learn knowledge from different domain teachers simultaneously,
we further present a novel domain decomposition module that learns to decompose the student model into two domain-related sub-parts. This is achieved by a novel domain-specific gate that learns to assign each filter to only one specific domain in a learnable way.
Extensive experiments demonstrate the effectiveness of our method. Codes and models are available at \textcolor{blue}{https://github.com/lovelyqian/ME-D2N\_for\_CDFSL}.
\end{abstract}



\begin{teaserfigure}\centering
  \vspace{-0.1in}
  \includegraphics[width=0.87\textwidth]{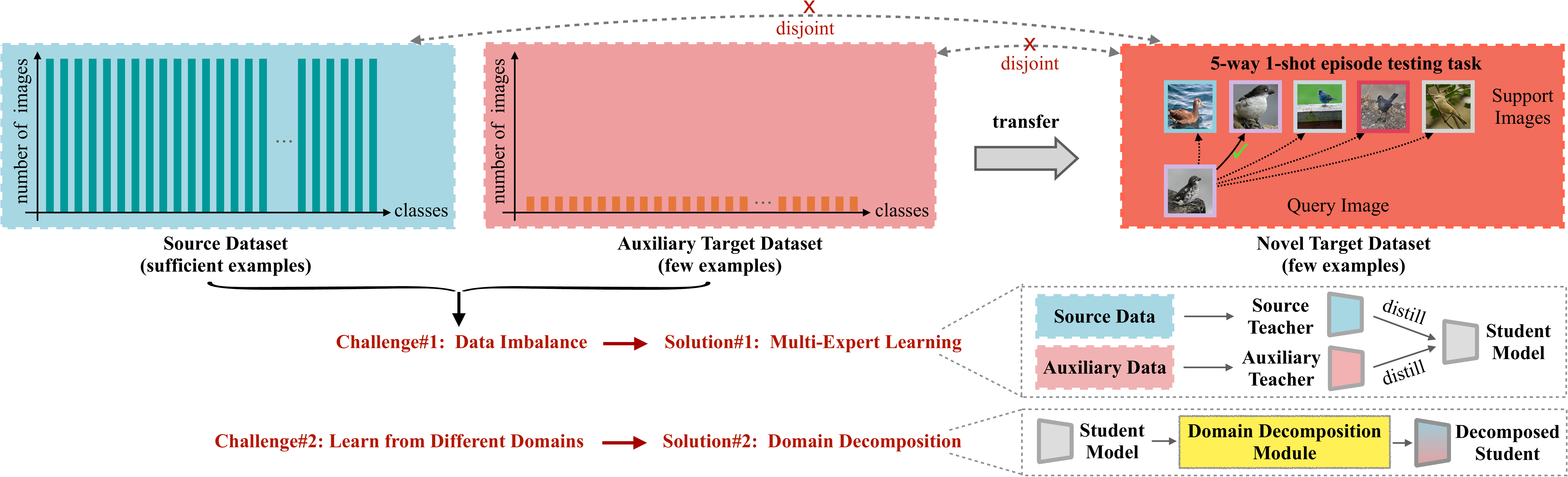}
  \vspace{-0.1in}
  \caption{
  Illustration of our motivation and solutions. We observe two core challenges: 1) serious data imbalance problem between the two training datasets; 2) network is required to learn from different domains simultaneously. Correspondingly, we have the following key solutions: 1) proposing the multi-expert learning that first learns two individual teacher models and then transfers the knowledge to a student model via knowledge distillation; 2) presenting a novel domain decomposition module that learns to decompose the network structure of student model into two domain-related sub-parts.}
  \label{fig:teaser}
\end{teaserfigure}
\maketitle

\section{Introduction}\label{sec:Intro}
FSL mainly aims at transferring knowledge from a source dataset to a novel target dataset with only one or few labeled examples. Generally, FSL assumes that the images of the source and target datasets belong to the same domain. However, such an ideal assumption may not be easy to be met in real-world multimedia applications. For example, as revealed in ~\cite{chen2019closer}, a model trained on the Imagenet~\cite{deng2009imagenet} which is mainly composed of massive and diverse natural images still fails to recognize the novel fine-grained birds.
To this end, CD-FSL which is dedicated to addressing the domain gap problem of FSL has invoked rising attention.

Recently, various settings of CD-FSL have been extensively studied in many previous methods~\cite{tseng2020cross, sun2020explanation, fu2022wave, phoo2020self, fu2021meta}. Most of them ~\cite{tseng2020cross, sun2020explanation, fu2022wave} use only the source domain images for training and pay efforts on improving the generalization ability of the FSL models. Though some achievements have been made, it is still hard to achieve very impressive performance due to the huge domain gap between the source and target datasets. Thus, some works~\cite{phoo2020self, fu2021meta} relax the most basic yet strict setting, and allow target data to be used during the training phase. More specifically, 
STARTUP~\cite{phoo2020self} proposes to make use of relative massive unlabeled target data, whilst Meta-FDMixup~\cite{fu2021meta} advocates utilizing few limited labeled target data.
Unfortunately, the massive unlabeled examples in 
the former one may still be not easy to be obtained in many real-world applications, such as the recognition of endangered wild animals and specific buildings.
By contrast, learning CD-FSL with few limited labeled target domain data, \textit{e.g., 5 images per class}, is more realistic. Thus, in this paper, we stick to the setting proposed in Meta-FDMixup~\cite{fu2021meta} to promote the learning process of models.

Formally, given a source domain dataset with enough examples and an auxiliary target domain dataset with only a few labeled examples, our goal is to learn a good FSL model taking these two sets as training data and achieve good results on the novel target data. Notably, as in Meta-FDMixup, our setting doesn't violate the basic FSL setting, as the class sets of the auxiliary target training data and the novel target testing data are disjoint from each other strictly. This ensures that none of the novel target categories will appear during the training stage.
Critically, as shown in Figure~\ref{fig:teaser}, we highlight that there are two key challenges: 1) The number of labeled examples for the source dataset and auxiliary target dataset are extremely unbalanced. Models learned on such unbalanced training data will be biased towards the source dataset while performing much worse on the target dataset. 
2) Since the source dataset and the auxiliary target belong to two distinct domains, 
it may be too difficult for a single model to learn knowledge from datasets with different domains simultaneously. Such challenges unfortunately have less been touched in previous works~\cite{fu2021meta}.

To address these challenges, this paper presents a novel \textbf{M}ulti-\textbf{E}xpert \textbf{D}omain \textbf{D}ecompositional \textbf{N}etwork (\textbf{ME-D2N}) for CD-FSL. 
Our key solutions are also illustrated in Figure~\ref{fig:teaser}.
Specifically, taking unbalanced datasets as training data will leads to the model biased problem~\cite{buda2018systematic, wang2020long}. That is, the learned model tends to perform well on the classes with more examples but has a performance degradation on the categories with fewer examples.
To tackle the data imbalance issue, we propose to build our model upon the multi-expert learning paradigm.
Concretely, rather than learning a model on the merged data of source and auxiliary target datasets directly, we train two teacher models on the source and the auxiliary dataset, respectively. Models trained in this way can be considered experts in their specialized domain avoiding being affected by training data of another domain.
Then, we transfer the knowledge from these two teachers to our student model. 
This is done by using the knowledge distillation technique which constrains the student model to produce consistent predictions with the teachers.
By distilling the \textit{individual knowledge} from both source and target teacher models, our student model picks up the ability to recognize both the source and auxiliary target images, avoiding learning from the \textit{unbalanced datasets}.
We take one step further: considering that forcing a unified model to learn from teachers of different domains may be nontrivial. Concretely, since each filter in the network needs to be responsible for extracting all domain features simultaneously, this vanilla learning method may limit the performance of the network.
A natural question is whether it is possible to decompose the student model into two parts -- one for learning from the source teacher and the another for the auxiliary target teacher?
Based on the above insights, a novel domain decomposition module which is also termed as D2N is proposed. Specifically, our D2N aims at building a one-to-one correspondence between the network filters and the domains. That is, each filter is only assigned to be activated by one specific domain. Technically, we achieve this by proposing a novel domain-specific gate that learns the activation state of filters for a specific domain dynamically. We insert the D2N into the feature extractor of the student model and make it learnable together with the model parameters.

We conduct extensive experiments on four different target datasets. Results well indicate that our multi-expert learning strategy helps address the data imbalance problem. Besides, our D2N further improves the performance of the student model showing the advantages of decomposing the student model into two domains.

\noindent\textbf{Contributions.} We summarize our contributions as below: 
1) For the first time, we introduce the multi-expert learning paradigm into the task of CD-FSL with few labeled target data to prevent the model from learning on unbalanced datasets directly. 
By learning from two teachers, we avoid our model being biased towards the source dataset with significantly more samples.
2) A novel domain decomposition module (D2N) is proposed to learn to decompose the model's filters into the source and target domain-specific parts.
The concept of domain decomposition has less been explored in previous work, especially for the task of CD-FSL. 
3) Extensive experiments conducted show the effectiveness of our modules and our proposed full model ME-D2N builds a new state of the art.

\section{Related Work}
\noindent \textbf{Cross-Domain Few-Shot Learning.}
Recent study~\cite{chen2019closer} finds that most of the existing FSL methods~\cite{snell2017prototypical, vinyals2016matching, sung2018learning, garcia2017few, finn2017model, rusu2018meta, rusu2018meta, li2020adversarial, TangLPT20, TANG2022108792, zhang2021curriculum, fu2020depth, zhang2022progressive, xie2022learning} that assume the source and target datasets belong to the same distribution fail to generalize to novel datasets with a domain gap. Thus, CD-FSL which aims at addressing FSL across different domains has risen increasing attentions~\cite{tseng2020cross, guo2020broader, sun2020explanation, wang2021cross, fu2022wave, phoo2020self, fu2021meta, islam2021dynamic, cai2021damsl}.
In this paper, these CD-FSL methods are categorized according to which kind of data are being used for training: 
1) CD-FSL with only source data~\cite{tseng2020cross, sun2020explanation, wang2021cross, fu2022wave}; 2) CD-FSL with unlabeled target data~\cite{phoo2020self, islam2021dynamic}; 3) CD-FSL with labeled target data~\cite{fu2021meta}.
Typically, CD-FSL with only source data is the most strict setting that demands model to recognize totally unseen target dataset without any target information.
Flagship works including FWT~\cite{tseng2020cross}, BSCD-FSL~\cite{guo2020broader}, LRP~\cite{sun2020explanation}, ATA~\cite{wang2021cross},  wave-SAN~\cite{fu2022wave}, RDC~\cite{li2021ranking}, NSAE~\cite{liang2021boosting}, and ConFT~\cite{das2021importance}.
Though many well-designed techniques e.g. readjusting the batch normalization~\cite{tseng2020cross}, augmenting the difficult of meta tasks~\cite{wang2021cross}, spanning style distributions~\cite{fu2022wave}, and even fine-tuning models using few target images during the testing stage~\cite{guo2020broader, liang2021boosting, das2021importance, li2021ranking}, the performances of them are still greatly limited due to the huge domain gap.
By contrast, STARTUP~\cite{phoo2020self} relaxes this strict setting and uses unlabeled target data for training.
Another choice that performs CD-FSL with few labeled target data is advocated by Meta-FDMixup~\cite{fu2021meta}, since obtaining extremely few labeled target data per class is relatively more realistic and can boost the model performance to a large extent.
Thus, in this paper, we mainly stick to the setting of CD-FSL with few labeled target data.

\noindent \textbf{Long-Tailed Recognition.}
This paper is also related to the long-tailed recognition from the perspective of tackling data imbalance problem.
In the literature, many attempts have been made to address the task of long-tailed recognition \cite{gao2022dynamic, zhu2022balanced}.
Main-stream methods include: 1) re-sampling based methods~\cite{buda2018systematic, byrd2019effect, he2009learning} which under-sample the head classes or over-sample the tail classes; 2) re-weighting based methods~\cite{huang2016learning, huang2019deep, wang2017learning, cai2022luna} which assign different weights for classes or instances; 3) two-stage fine-tuning based methods~\cite{cao2019learning, kang2019decoupling, zhou2020bbn} which train the representation and classifier separately.
Another line of work~\cite{wang2020long, xiang2020learning, cai2021ace} trains multiple experts for the head, medium, and tail classes respectively which is the most related work to us.
However, to the best of our knowledge, it is the first time that multi-expert learning is used for the CD-FSL task. Our method trains two teacher models of different domains and transfers the knowledge into the student model.
In addition, besides the unbalanced training sets, what intrinsically distinguishes our work from these methods is that we tackle datasets of different domains. The idea of decomposing the student model into different domains has never been explored in these works.

\noindent \textbf{Decomposition of Network Filters.}
Generally, as studied in ~\cite{zhang2018interpretable, bau2017network}, the filters of a normal CNN tend to extract mixed features of the input data. Such entangled filters inevitably lead to some unexpected problems, including limiting the representational capability of the network and increasing the uninterpretability.
Subsequently, some methods~\cite{qiu2018dcfnet, ioannou2015training, ruan2020edp, li2020group} explore decomposing the filters for making a more efficient and compressed network. Other works including ~\cite{liang2020training, zhang2018interpretable, shen2021interpretable, chen2022graph} decompose the network filters for more interpretable networks via assigning filters dynamically. Generally, we are similar but fundamentally different from methods of this type. They learn the correspondence between filters and the ``classes'' or ``objects'' to explain the activation of the model, while our motivation is to decompose the student model into two ``domains'' so that we can learn from teachers of different domains.
Another work that may also be related to us is domain-aware dynamic network~\cite{zhang2019domain} which learns different weights for different domains.
However, using soft weights for readjusting the activation of the network essentially can not be seen as a decomposition.

\begin{figure*}
  \centering
  \includegraphics[width=0.9\textwidth]{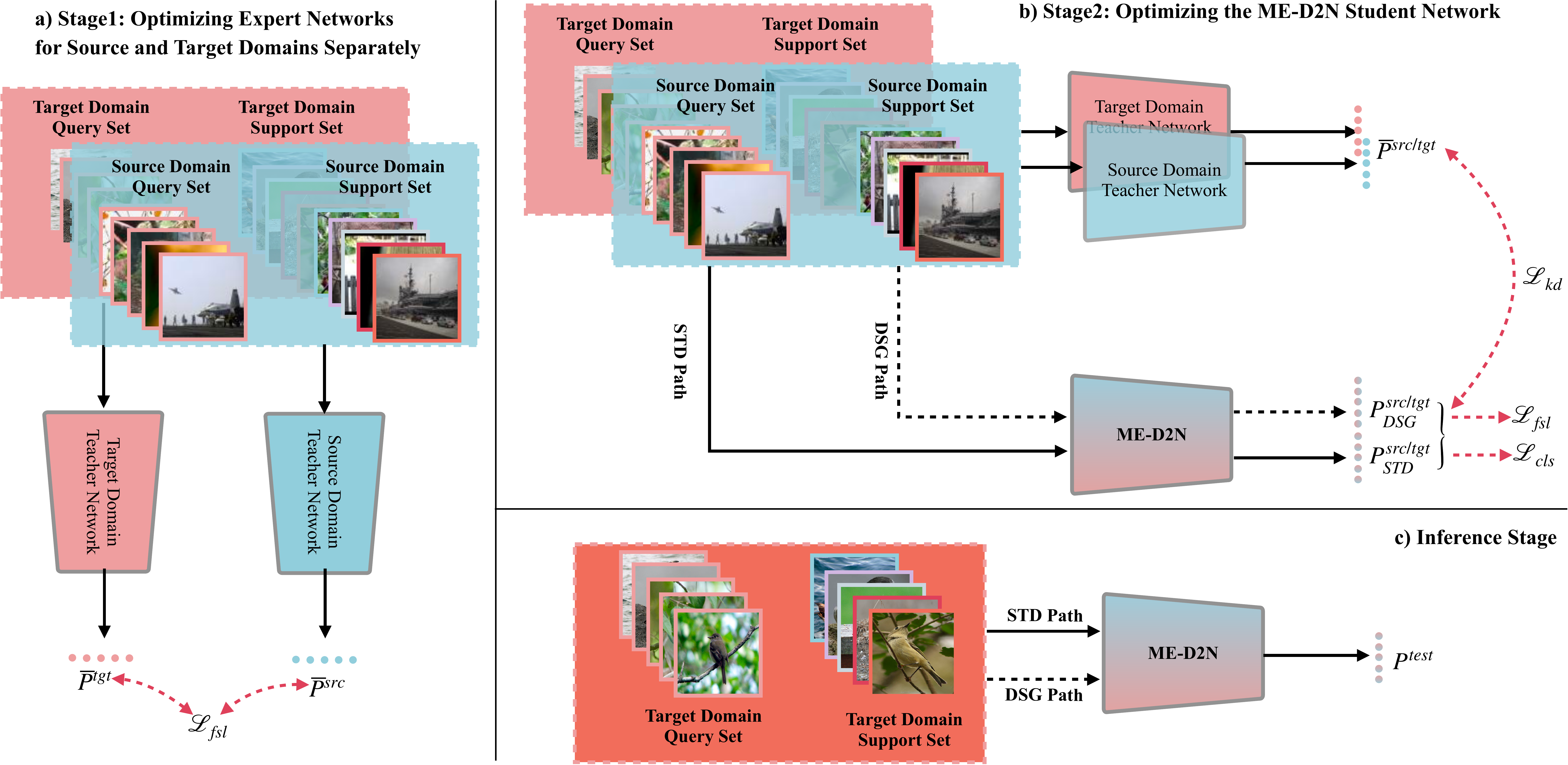}
  \caption{Our method contains two training stages: a) optimizing the experts network for the source and target domains; b) optimizing the ME-D2N student network by distilling knowledge from both source and target domain experts. In the inference stage, only the ME-D2N network is used for prediction.}
  \vspace{-0.15in}
  \label{fig:NetworkStructure}
\end{figure*}

\section{Method}\label{sec:method}
\noindent \textbf{Problem Definition.}
For the CD-FSL with few labeled target data, we have two training sets: source training dataset $D_{src} = \left \{x_{src}, y_{src} \right \}$ and the auxiliary target dataset $D_{tgt} = \left \{ x_{tgt}, y_{tgt} \right \}$.
The model trained on $D_{src}$ and $D_{tgt}$ is evaluated on the novel target testing dataset $D_{test} = \left \{x_{test}, y_{test} \right \}$. The $x$ represents the image examples and the $y$ denotes the corresponding labels of images. Note that all the classes contained in $D_{src}$, $D_{tgt}$, and $D_{test}$ are \textbf{disjoint} from each other and there is a domain gap between the source dataset $D_{src}$ and the target datasets $D_{tgt}, D_{test}$.

We construct meta-learning tasks which also known as the \textbf{$N$-way $K$-shot episodes} to train and test our model. Typically, an episode contains a support set $S = \left \{ x_{i},y_{i} \right \}_{i}^{N\times K}$ and a query set $Q = \left \{ x_{i},y_{i} \right \}_{i}^{N\times M}$. 
$N$-way $K$-shot means that $N$ categories are sampled. Each category of $S$ and $Q$ contains $K$ labeled examples and $M$ testing images, respectively.
The images in $Q$ are classified according to the given $S$.
We use the $\left\{S_{src}, Q_{src}\right\}$, $\left\{S_{tgt}, Q_{tgt}\right\}$, and $\left\{S_{test}, Q_{test}\right\}$ to denote episodes sampled from $D_{src}$, $D_{tgt}$, and $D_{test}$, respectively.

\noindent \textbf{Method Overview.}
The overall illustration of our method is given in Figure~\ref{fig:NetworkStructure}. We mainly have two training stages: a) optimizing the source and target domain teacher networks (\textbf{St-Net} \& \textbf{Tt-Net}) separately; b) optimizing the multi-expert domain decompositional student network (\textbf{ME-D2N}) by distilling the knowledge from St-Net and Tt-Net. The St-Net and Tt-Net are composed of an embedding module $E$ and an FSL classifier $G$. 
Besides these two basic modules $E, G$, ME-D2N also contains a novel domain decomposition module $D2N$ and two global classifiers $f_{src}$,  $f_{tgt}$ which classify the input images into the global source and target categories. Note that $f_{src}$,  $f_{tgt}$ are only used during the training phase.
As for the object functions, the St-Net and Tt-Net are optimized using the FSL classification loss $\mathcal{L}_{fsl}$ alone. The ME-D2N is optimized by the $\mathcal{L}_{fsl}$, the knowledge distillation loss $\mathcal{L}_{kd}$, and the global classification loss $\mathcal{L}_{cls}$ simultaneously. Note that we use the ``STD path'' and the ``DSG path'' in the figure to denote the standard forward path and the domain-specific gate forward path which is guided by the D2N, respectively. During testing, the ME-D2N is utilized to obtain the predictions for the novel target episodes.

\subsection{Learning the Teacher Networks}
As shown in Figure~\ref{fig:NetworkStructure}.a, we first train our two teacher networks (St-Net \& Tt-Net) using episodes sampled from the source training dataset $D_{src}$ and the auxiliary target dataset $D_{tgt}$, respectively. These trained teachers are considered experts in the corresponding domain to guide the subsequent training of the ME-D2N student network. Both the network structure and training process of St-Net and Tt-Net are exactly the same. Here we take the St-Net as an example to introduce the learning details of the teacher network.
For each training iteration, we randomly sample a source episode $\left\{S_{src}, Q_{src}\right\}$ from $D_{src}$ as input, and feed it into the embedding module $E$ of St-Net to obtain the feature representations of the $S_{src}$ and $Q_{src}$. After that, the FSL classifier module $G$ of St-Net is used to predict the class categories of $Q_{src}$ according to the $S_{src}$ resulting in the FSL prediction scores $\overline{P}^{src}$.
Note that to prevent the model from learning the correspondence between the inputs images to its global labels $y^{src}$, meta-learning refines the labels of these $N$ classes as $y_{fsl}^{src} \in (0, 1, ..., N-1)$.
By calculating the cross entropy loss between the predictions and its corresponding FSL ground truth $y_{fsl}^{src}$ as follows, we obtain its FSL classification loss $\mathcal{L}^{src}_{fsl}$.

\begin{equation}\label{equa:fsl}
    \mathcal{L}_{fsl}^{src} = CE \left( \overline{P}^{src},y_{fsl}^{src} \right )
\end{equation}

where $CE$ indicates the cross entropy loss. In the same way, for the Tt-Net, we get the FSL prediction scores $\overline{P}^{tgt}$ and the FSL classification loss $\mathcal{L}^{tgt}_{fsl}$. The $\mathcal{L}^{src}_{fsl}$ and the $\mathcal{L}^{tgt}_{fsl}$ are finally used for optimizing the Tt-Net and St-Net, respectively.

\subsection{Domain Decomposition Module}\label{sec:DSG}
Given two experts St-Net and Tt-Net, a direct and commonly used solution for training the student model is distilling knowledge from these two teachers at the same time. However, as we stated in Sec.~\ref{sec:Intro}, one key challenge of this task lies in the available training sets $D_{src}$ and $D_{tgt}$ belong to different domains. To that end, the learned teacher models will be biased towards their training domains.
It may be difficult for a unified student model to learn knowledge from two teachers of different domains. Thus, our D2N learns to decompose the student model into the source-specific part and target-specific part. Overall, the decomposition is achieved by a novel domain-specific gate (DSG) that learns to assign each filter to only one specific domain dynamically.

\begin{figure}[htb]
  \centering
  \includegraphics[width=0.9\linewidth]{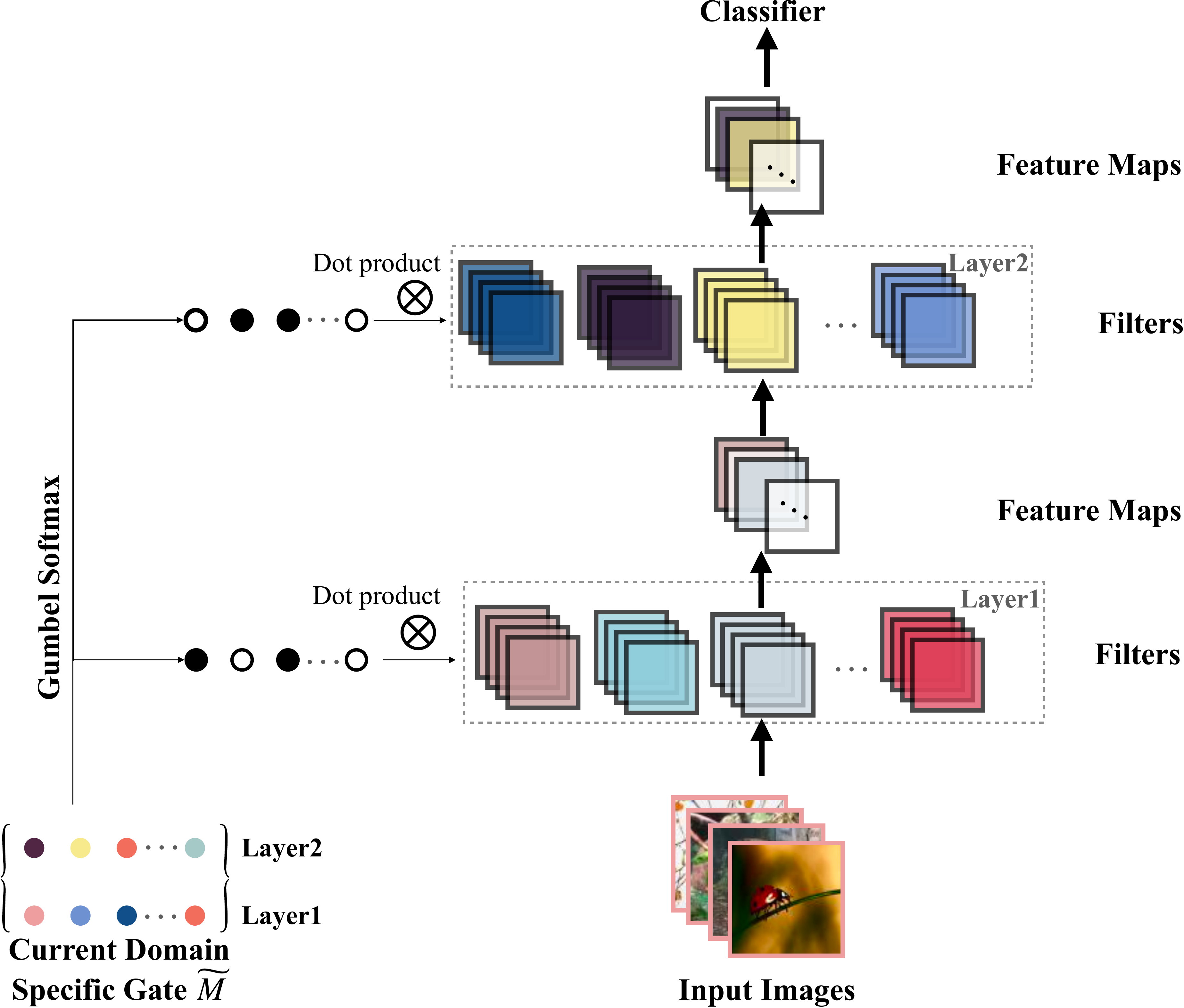}
  \caption{
 The illustration of our domain decomposition module (D2N). D2N learns a domain-specific gate matrix $\widetilde{M}$ to control the activation states of the filters. The Gumbel softmax is utilized to binarize the soft matrix.}
  \vspace{-0.25in}
  \label{fig:NetworkDecompose}
\end{figure}

As shown in Figure~\ref{fig:NetworkDecompose}, we first randomly initialize the domain-specific gate matrix $\widetilde{M}$. The number of elements $\widetilde{M}$ is consistent with the number of filters that need to be decomposed in the network. The element in $\widetilde{M}$ can be seen as the probability that the corresponding filter belongs to one specific domain. Typically, we use the $\widetilde{M}$ to denote the gate matrix for the source domain, and the gate matrix for the target domain can be easily obtained by $1-\widetilde{M}$.
While the soft gate is not enough to achieve the real decomposition, which does not meet our ideal expectation of assigning a filter to only a domain. Thus, the Gumbel softmax~\cite{jang2016categorical} which generates the discrete data from a soft categorical distribution is introduced to transform the soft $\widetilde{M}$ into the hard one $M$ (denoted as the black and white dots).
With the $M$, the DSG forward path only activates the filters when the gates for them equal 1 thus establishing a one-to-one correspondence between filters and domains.

\subsection{Learning the ME-D2N Student Network}
As shown in Figure~\ref{fig:NetworkStructure}.b, to learn knowledge from both the source and target domain teachers, two episodes $\left\{S_{src}, Q_{src}\right\}$, $\left\{S_{tgt}, Q_{tgt}\right\}$ are randomly sampled as the input for each training iteration. These source and target episodes are fed into the ME-D2N through two different forward paths. One is the standard (STD) path and the other is the DSG path as introduced in Sec.\ref{sec:DSG}. Regardless of whether the forward is domain decomposed or not, all the other details e.g. input data and loss functions are the same for these two paths. Thus, for convenience, we first introduce the learning and optimization process of the DSG path as an example. For each input episode e.g. the source episode $\left\{S_{src}, Q_{src}\right\}$, we feed it into the embedding module and the FSL classifier module consequently obtaining its FSL predictions $P_{DSG}^{src}$. Based on the $P_{DSG}^{src}$, a total of three sub-tasks are performed resulting in three different losses. Firstly and most importantly, the knowledge distillation is performed to keep the ability of ME-D2N with that of the teacher model.  Specifically, the same $\left\{S_{src}, Q_{src}\right\}$ is fed into the trained St-Net obtaining the FSL predictions $\overline{P}^{src}$. After that, the Kullback-Leibler divergence loss is used to constrain the consistency between the $P_{DSG}^{src}$ and $\overline{P}^{src}$. Thus, the knowledge distillation loss $\mathcal{L}_{DSG_{\_kd}}^{src}$ is expressed as:

\begin{equation}
    \mathcal{L}_{DSG_{\_kd}}^{src} = KL \left( P_{DSG}^{src},\overline{P}^{src}\right )
\end{equation}

where the $KL$ means the Kullback-Leibler divergence loss. Secondly, by calculating the cross entropy loss between the predictions $P_{DSG}^{src}$ and its FSL ground truth as defined in Equa.~\ref{equa:fsl}, we obtain its FSL classification loss $\mathcal{L}_{DSG_{\_fsl}}^{src}$. Thirdly, we also use the global classifier $f_{src}$ to classify the input images into the its global class categories. This generates the global classification loss $\mathcal{L}_{DSG_{\_cls}}^{src}$. In the same way, we obtain the target knowledge distillation loss $\mathcal{L}_{DSG_{\_kd}}^{tgt}$, the target FSL classification loss $\mathcal{L}_{DSG_{\_fsl}}^{tgt}$, and the target global classification loss $\mathcal{L}_{DSG_{\_cls}}^{tgt}$. Formally, we have:

\begin{gather}
   \mathcal{L}_{DSG_{\_kd}} = \lambda_{1} \mathcal{L}_{DSG_{\_kd}}^{src} + (1-\lambda_{1}) \mathcal{L}_{DSG_{\_kd}}^{tgt} \\
   \mathcal{L}_{DSG_{\_fsl}} = \lambda_{1} \mathcal{L}_{DSG_{\_fsl}}^{src} + (1-\lambda_{1}) \mathcal{L}_{DSG_{\_fsl}}^{tgt} 
   \\
   \mathcal{L}_{DSG_{\_cls}} = \lambda_{1} \mathcal{L}_{DSG_{\_cls}}^{src} + (1-\lambda_{1}) \mathcal{L}_{DSG_{\_cls}}^{tgt} 
   \\
   \mathcal{L}_{DSG} =  \mathcal{L}_{DSG_{\_kd}} + \lambda_{2}\mathcal{L}_{DSG_{\_fsl}} + \lambda_{3}\mathcal{L}_{DSG_{\_cls}}
\end{gather}

Similarly, we obtain the loss for the STD path as $\mathcal{L}_{STD}$. The final loss for our ME-D2N is defined as follows:
\begin{equation}
    \mathcal{L} = \mathcal{L}_{DSG} + \lambda_{4}\mathcal{L}_{STD}
\end{equation}
The $\lambda_{1}$, $\lambda_{2}$, $\lambda_{3}$, and $\lambda_{4}$ are four hyper-parameters.

\noindent\textbf{Inference Stage.}
During the inference stage, as shown in Figure~\ref{fig:NetworkStructure}.c, only the ME-D2N is used for generating the predictions. Given a novel target testing episode $\left\{S_{test}, Q_{test}\right\}$, we forward it into the ME-D2N using both the DSG and STD paths resulting in two predictions $P_{DSG}^{test}$ and $P_{STD}^{test}$. Note that since the Gumbel softmax will cause inconsistencies in each forward process, unlike the training process, we directly binarize the gate matrix $\Tilde{M}$ by choosing the domain with a higher value.
The final prediction $P^{test}$ takes the mean of these two paths: 
\begin{equation}
    P^{test} = \frac{1}{2}(P_{DSG}^{test} + P_{STD}^{test})
\end{equation}

\section{Experiments}
\subsection{Setup}\label{sec:setup}
\noindent\textbf{Datasets.}
Totally five datasets are used to validate the effectiveness of our method. Concretely, mini-Imagenet~\cite{ravi2016optimization} works as the source dataset. CUB~\cite{wah2011caltech}, Cars~\cite{krause20133d}, Places~\cite{zhou2017places}, and Plantae~\cite{van2018inaturalist} serve as the target datasets, respectively.
As for the splits of $D_{src}$, $D_{tgt}$, and $D_{test}$, we strictly follow the meta-FDMixup~\cite{fu2021meta}. Specifically, each class category in $D_{tgt}$ has only 5 labeled examples.

\noindent\textbf{Network Modules.}
Resnet-10~\cite{he2016deep} is used as the embedding module $E$. GNN~\cite{garcia2017few} is selected as the FSL classifier $G$. Totally same basic modules $E$ and $G$ with previous CD-FSL methods ensures the fair comparisons.
Besides, same to wave-SAN~\cite{fu2022wave}, we divide the embedding module $E$ into four blocks and decompose the filters of the last two blocks. The global classifiers $f_{src}$ and $f_{tgt}$ are represented by a fully connected layer, respectively.

\noindent\textbf{Implementation Details.}
We conduct our experiments in the form of 5-way 1-shot and 5-way 5-shot meta tasks. 
For training, we train the Tt-Net 100 and 400 epochs for 5-way 1-shot and 5-way 5-shot, respectively. The training of both the St-Net and ME-D2N takes 400 epochs. 
The Adam with the initial learning rate of 0.001 is uniformly used as the optimizer.
Besides, as a common practice in CD-FSL~\cite{tseng2020cross, fu2021meta}, the $E$ pre-trained on the source dataset $D_{src}$ with the standard classification tasks is used to warm start the training of St-Net, Tt-Net, and ME-D2N.
As for the testing stage, the average accuracy of 1000 episodes randomly sampled from the novel target set $D_{test}$ is reported.
Without searching the best hyper-parameters for different target dataset, we uniformly set the $\lambda_1$,  $\lambda_2$, $\lambda_3$, and  $\lambda_4$ as 0.2, 0.05, 0.05, and 0.2, respectively.

\subsection{Baselines \& Competitors}
We introduce our baselines and competitors as below. 
\textbf{1) Typical FSL Methods.} Totally three flagship methods in the FSL community including MatchingNet~\cite{vinyals2016matching}, RelationNet~\cite{sung2018learning}, and GNN~\cite{garcia2017few} are introduced. 
These methods demonstrate how the typical FSL methods perform under the CD-FSL setting.
\textbf{2) Several Baselines.} Three baselines namely ``St-Net'', ``Tt-Net'', and ``M-base'' are included for comparisons. Concretely, as stated in Sec.~\ref{sec:method}, ``St-Net'' and ``Tt-Net'' represent our two teacher models. 
``M-base'' is obtained by training model on the merged data of source and auxiliary datasets.
The network architecture of ``M-base'' is exactly the same as that of ``St-Net'' and ``Tt-Net''. 
These three baselines play an important role in analyzing the effectiveness of our method.
\textbf{3) CD-FSL Methods.} Since the setting of CD-FSL with few labeled target data is proposed very recently, meta-FDMixup~\cite{fu2021meta} is the only competitor that can be compared directly. To better illustrate the competitiveness of our method, we further adapt several other CD-FSL methods including FWT~\cite{tseng2020cross} which improves the generalization ability of the model by feature-wise transformation, 
LRP~\cite{sun2020explanation} which utilizes the results of the explanation model to guide the learning process, ATA~\cite{wang2021cross} which augments the meta tasks via adversarial attack, and wave-SAN~\cite{fu2022wave} which tackles the domain gap from the perspective of augmenting the style distributions of the source dataset. These methods are initially designed for the most strict CD-FSL setting. 
The adaption is achieved by training the models using the merged training data as used for ``M-base''. To that end, we obtain the competitors ``M-FWT'', ``M-LRP'', ``M-ATA'', and ``M-waveSAN''.

	\begin{table*}[t]
		\centering
		\begin{tabular}{llcccccc}
			\hline
			\textbf{5-way} & \textbf{1-shot} & \textbf{$D_{tgt}$} & \textbf{CUB} &  \textbf{Cars} & \textbf{Places} & \textbf{Plantae} & \textbf{Avg.}\\
			\hline
			FSL 
				& MatchingNet~\cite{vinyals2016matching} & - &  35.89 \mypm 0.51 & 30.77 \mypm 0.47 & 49.86 \mypm 0.79 & 32.70 \mypm 0.60 & 37.31 \\
				
			    & RelationNet~\cite{sung2018learning} & - & 42.44 \mypm 0.77 &  29.11 \mypm 0.60 & 48.64 \mypm 0.85 &  33.17 \mypm 0.64 & 38.34\\
			    
			     & GNN~\cite{garcia2017few} &  - &  45.69 \mypm 0.68 & 31.79 \mypm 0.51 & 53.10 \mypm 0.80 & 35.60 \mypm 0.56 & 41.54 \\
	
			\hline
			Baselines 
			& St-Net & - & 46.10 \mypm 0.68  &	31.05 \mypm 0.54  &	54.22 \mypm 0.81 	& 37.11 \mypm 0.60 & 42.12 \\
			& Tt-Net & \Checkmark & 52.35 \mypm 0.79 	& 39.16 \mypm 0.65  & 49.49 \mypm 0.78  & 44.54 \mypm 0.75 & 46.39 \\
			
			& M-base  & \Checkmark &  57.65 \mypm 0.80 & 46.03 \mypm 0.72             &  55.70 \mypm 0.79 & 48.25 \mypm 0.74 & 51.91 \\
			\hline
			
			CD-FSL 
			& FWT~\cite{tseng2020cross}  & - & 47.47\mypm0.75 & 31.61\mypm0.53 & 55.77\mypm0.79 & 35.95\mypm0.58 & 42.70\\
			
			& LRP~\cite{sun2020explanation}  & - & 48.29\mypm0.51 & 32.78\mypm0.39 & 54.83\mypm0.56  & 37.49\mypm0.43 & 43.35 \\
				
			& ATA~\cite{wang2021cross} & - & 45.00\mypm0.50 & 33.61\mypm0.40 & 53.57\mypm0.50 & 34.42\mypm0.40 & 41.65\\
			
			& wave-SAN~\cite{fu2022wave} & - &   50.25\mypm0.74 & 
			33.55\mypm0.61 & 
		    57.75\mypm0.82 & 40.71\mypm0.66 & 45.57 \\
			\cline{2-8}
			&	M-FWT~\cite{tseng2020cross} &  \Checkmark & 61.16  \mypm  0.81 & 49.01  \mypm  0.76  & 57.89  \mypm  0.82  &  50.49  \mypm  0.81 & 54.64 \\
			& M-LRP~\cite{sun2020explanation}$\dagger$ & \Checkmark & 59.23 \mypm 0.58 &
	46.88 \mypm 0.53 &
	57.92 \mypm 0.58 &
	49.11 \mypm 0.54 & 53.29 \\
			& M-ATA~\cite{wang2021cross}$\dagger$ & \Checkmark & 57.73 \mypm 0.57 &
	45.19 \mypm 0.49 &
	55.39 \mypm 0.55 & 
	48.07 \mypm 0.52 & 51.60
 \\
			& M-waveSAN~\cite{fu2022wave} $\dagger$ &\Checkmark &  63.59 \mypm 0.85  &	50.06 \mypm 0.76 &	59.89 \mypm 0.86 &	51.99 \mypm 0.81 & 56.38\\
			& meta-FDMixup~\cite{fu2021meta} & \Checkmark &  63.24 \mypm  0.82  & \textbf{51.31 \mypm  0.83 } & 58.22 \mypm  0.82 & 51.03 \mypm  0.81 & 55.95\\
			\hline
			\textbf{Ours} & \textbf{ME-D2N} & \Checkmark & \textbf{65.05 \mypm 0.83}  & 49.53 \mypm 0.79  & \textbf{60.36 \mypm 0.86}  & \textbf{52.89 \mypm 0.83} & \textbf{56.96} \\
			\hline
			\hline
			\textbf{5-way} & \textbf{5-shot} & \textbf{$D_{tgt}$} &  \textbf{CUB} &  \textbf{Cars} & \textbf{Places} & \textbf{Plantae} & \textbf{Avg.}\\
			\hline
			FSL 
				& MatchingNet~\cite{vinyals2016matching} & - & 51.37 \mypm 0.77 & 38.99 \mypm 0.64 & 63.16 \mypm 0.77  & 46.53 \mypm 0.68 & 50.01\\
				
			    & RelationNet~\cite{sung2018learning} & - & 57.77 \mypm 0.69 & 37.33 \mypm 0.68 & 63.32 \mypm 0.76 & 44.00 \mypm 0.60 & 50.61\\
			    
			     & GNN~\cite{garcia2017few} &  - &  62.25 \mypm 0.65 &  44.28 \mypm 0.63 & 70.84 \mypm 0.65 & 52.53 \mypm 0.59 & 57.48\\
			\hline
			Baselines  
			& St-Net & - &  66.89 \mypm 0.66 	& 46.26 \mypm 0.67  &	72.87 \mypm 0.67  	& 55.13 \mypm 0.66 & 60.29\\
		
			& Tt-Net & \Checkmark &  64.72 \mypm 0.69  	& 52.32 \mypm 0.69  & 69.37 \mypm 0.68  & 59.23 \mypm 0.70 & 61.41\\
			
			& M-base    & \Checkmark &   78.08 \mypm 0.60 &  63.27 \mypm 0.70 &  75.90 \mypm 0.67 & 66.69 \mypm 0.68 & 70.99\\
			\hline
			CD-FSL
			&FWT~\cite{tseng2020cross}   & - 
				& 66.98\mypm0.68 
				& 44.90\mypm0.64 
				& 73.94\mypm0.67 
				& 53.85\mypm0.62 & 59.92\\
				
				&LRP~\cite{sun2020explanation}  & - 
				& 64.44\mypm0.48 
				& 46.20\mypm0.46 
				& 74.45\mypm0.47 
				& 54.46\mypm0.46 & 59.89\\
				
				&ATA~\cite{wang2021cross}  & - 
				& 66.22\mypm0.50 
				& 49.14\mypm0.40 
				& 75.48\mypm0.40 
				& 52.69\mypm0.40 & 60.88\\
			
			& wave-SAN~\cite{fu2022wave} & - & 
			70.31\mypm0.67 & 
			46.11\mypm0.66 & 
			76.88\mypm0.63  &  57.72\mypm0.64 & 62.76\\

			\cline{2-8}
			& M-FWT~\cite{tseng2020cross}  & \Checkmark &  79.14  \mypm  0.62  &  65.42  \mypm  0.70  & 78.59  \mypm  0.60  &  68.26  \mypm  0.68 & 72.85\\
			& M-LRP~\cite{sun2020explanation} $\dagger$ & \Checkmark &  77.07 \mypm 0.44 &
	64.38 \mypm 0.48 &
	77.73 \mypm 0.45 &
	67.90 \mypm 0.47 & 71.77
\\
			& M-ATA~\cite{wang2021cross} $\dagger$ & \Checkmark &  73.96 \mypm 0.46 &
	68.58 \mypm 0.45 &
	76.73 \mypm 0.42 &
	66.45 \mypm 0.46 & 71.43
	\\
			& M-waveSAN~\cite{fu2022wave} $\dagger$ & \Checkmark &  82.29 \mypm 0.58 & 66.93 \mypm 0.71 & 80.01 \mypm 0.60 & 71.27 \mypm 0.70 & 75.13 \\
			& meta-FDMixup~\cite{fu2021meta} & \Checkmark &  79.46 \mypm 0.63  & 66.52 \mypm 0.70 & 78.92 \mypm 0.63 & 69.22 \mypm 0.65 & 73.53\\
			\hline
			\textbf{Ours} & \textbf{ME-D2N} & \Checkmark & \textbf{83.17 \mypm 0.56}  & 	\textbf{69.17 \mypm 0.68}  & \textbf{80.45 \mypm 0.62}  & \textbf{72.87 \mypm 0.67} & \textbf{76.42} \\
			\hline
		\end{tabular}
		\caption{The 5-way 1(5)-shot classification results (\%) on four novel target datasets. ``Avg.'' is short for ``Average Accuracy''. The checkmark indicates whether the auxiliary target data $D_{tgt}$ is used for training. Notation $\dagger$ denotes that we adapt the methods into our setting. In most cases, our ME-D2N outperforms all the FSL, baselines, and CD-FSL competitors.}
				\label{tab:main-target}
				\vspace{-0.2in}

	\end{table*}

\subsection{Main Results}

\begin{table*}[t!]
		\centering
		\begin{tabular}{lccccc}
			\hline
			\textbf{Methos} &  \textbf{mini-Img $|$ CUB} & \textbf{mini-Img $|$ Cars} & \textbf{mini-Img $|$ Places} & \textbf{mini-Img $|$ Plantae} & \textbf{Avg.} \\
			\hline
	
			St-Net & 81.36 \mypm 0.57  & 81.36 \mypm 0.57  & 81.36 \mypm 0.57  & 81.36 \mypm 0.57  & 81.36\\

			Tt-Net & 52.02 \mypm 0.66  & 53.91 \mypm 0.70  & 64.77 \mypm 0.71 & 51.63 \mypm 0.69  & 55.58 \\
			
			M-base & 78.94  \mypm 0.58  & 80.75   \mypm 0.55  & 79.99   \mypm 0.58 & 80.51   \mypm 0.55 & 80.05\\
			
			\hline
			
			M-FWT & 81.88 \mypm 0.57 & 80.89 \mypm 0.58 & 81.32   \mypm 0.56  & 82.28   \mypm 0.55 & 81.59\\
			M-LRP~\cite{sun2020explanation} $\dagger$& 80.84 \mypm 0.40 &
81.15 \mypm 0.41 &
81.07 \mypm 0.40 &
81.51 \mypm 0.38 & 81.14
 \\
			
			M-ATA~\cite{wang2021cross} $\dagger$& 77.84 \mypm 0.39 &
78.49 \mypm 0.40 &
77.57 \mypm 0.39 &
78.39 \mypm 0.40 & 78.07
\\
			
			M-waveSAN~\cite{fu2022wave} $\dagger$& 83.08 \mypm 0.56 & 82.96 \mypm 0.57 & 82.79 \mypm 0.56 & 83.20 \mypm 0.58 & 83.01\\
		
		    meta-FDMixup~\cite{fu2021meta} &  82.29 \mypm 0.57  & 81.00   \mypm 0.58 & 81.37   \mypm 0.56   & 79.64 \mypm 0.59  & 81.08\\
		    \hline
		    \textbf{ME-D2N} & \textbf{84.10 \mypm 0.53}  &	\textbf{83.89 \mypm 0.53} 	& \textbf{85.45 \mypm 0.53}  & \textbf{83.74 \mypm 0.56} & \textbf{84.30}\\
			\hline
			
		\end{tabular}
		\caption{
		The 5-way 5-shot results  (\%) on the testing set of mini-Imagenet (abbreviated as mini-Img). 
		``mini-Img | target set" indicates the model is trained on which target dataset. 
		Note that ``St-Net'' doesn't need any target data thus its results on four target datasets are the same.
		``Avg.'' is short for ``Average Accuracy''. Our ME-D2N shows clear advantages over other competitors.}	
		\vspace{-0.25in}
		\label{tab:main-source}
	\end{table*}

	\begin{table*}[t]
		\centering
		\begin{tabular}{cccccccc}
			\hline
			\textbf{Method} & \textbf{ME} &  \textbf{D2N} & \textbf{CUB} & \textbf{Cars} & \textbf{Places} & \textbf{Plantae} & \textbf{Avg.}\\
			\hline
			
			M-base & - & - &  78.08 \mypm 0.60 &  63.27 \mypm 0.70 &  75.90 \mypm 0.67 & 66.69 \mypm 0.68 & 70.99 \\
			ME & \Checkmark & - & 82.22 \mypm 0.56 &	66.59 \mypm 0.73  &	79.63 \mypm 0.64 	& 71.30 \mypm 0.68 & 74.94\\
			ME + D2N (ours) & \Checkmark & \Checkmark & \textbf{83.17 \mypm 0.56}  & \textbf{69.17 \mypm 0.68} & \textbf{80.45 \mypm 0.62} & \textbf{72.87 \mypm 0.67} & \textbf{76.42}\\
			\hline
			
		\end{tabular}
		\caption{The effectiveness of our main technical contributions -- multi-expert learning (abbreviated as ME) and domain decomposition module (D2N) are shown. Experiments were conducted under the 5-way 5-shot setting.}
		\vspace{-0.15in}
		\label{tab:modules}
	\end{table*}

\noindent\textbf{Main Results on Target Datasets.}
The comparison results of our ME-D2N against all the typical FSL methods, baselines, CD-FSL with or without auxiliary target training data $D_{tgt}$ are given in Table~\ref{tab:main-target}.   
What can be apparently seen from the results is that our ME-D2N achieves the best results beating all the baselines and competitors in most cases. Specifically, under the 5-way 5-shot setting, we achieve 83.17\%, 69.17\%, 80.45\%, and 72.87\% on the cub, cars, places, and plantae, respectively. Compared with the GNN, our ME-D2N has an average improvement of 15.42\% and 18.94\% on 5-way 1-shot and 5-shot tasks. Such a performance growth is contributed by both the use of auxiliary target data and the effectiveness of our technical solutions.
Besides, there are some other points worth mentioning. 1) Firstly, by comparing the results of ``M-base'' with that of FWT, LRP, ATA, and wave-SAN, we observe that by merging the source and target datasets together directly as the training data, the ``M-base'' shows advantages over these methods that are carefully designed for CD-FSL. Another observation is that after adapting these methods to our setting e.g. adapting the FWT to M-FWT, an obvious improvement can be found. 
These two phenomenons together show the superiority of introducing the few auxiliary target data and well explain why we stick to this setting;
2) Though the M-base has achieved relatively good performance, our ME-D2N still outperforms it by a large margin. This basically shows that we further address the data imbalance issue thus the knowledge of the training data is utilized to a larger extent; 3) We also notice that the performances of our teacher models St-Net and Tt-Net are not so good. Take the Tt-Net as an example, it has only an average accuracy of 46.39\% and 61.41\% on 1-shot and 5-shot settings. However, based on these ``ordinary'' teachers, our final ME-D2N still achieves very high results. This indicates that the knowledge learning process of our student model is effective and partially shows that our domain decomposition module makes the student model not limited to the performance of teachers.

As for the CD-FSL competitors with auxiliary target data, generally, the M-ATA performs worst, then follows the M-LRP and M-FWT. Meta-FDMixup and M-waveSAN are the most competitive methods. 
The relatively good performance of meta-FDMixup is not easy to understand since it is purely proposed for this setting.
Among these competitors, the M-waveSAN performs best with very competitive results. This shows that the idea of augmenting styles is also a helpful solution to narrow the domain gap. But, generally, M-waveSAN is still inferior to our method.

\noindent\textbf{Main Results on Source Dataset.}
To further test the performance of our method on the original source domain, we compare it against the baselines and CD-FSL competitors. The 5-way 5-shot results on the testing set of the mini-Imagenet (disjoint from $D_{src}$), abbreviated as mini-Img, are given in Table~\ref{tab:main-source}. Since our setting is target dataset specific, the results are reported in the form of ``mini-Img | target set''. Notably, the St-Net is trained using only source data. Thus, the results of St-Net on four target datasets are the same. 

We mainly have the following observations. 1) Our ME-D2N outperforms all the three baseline methods and five CD-FSL competitors achieving an average accuracy of 84.30\%. This demonstrates that our model keeps the best ability of recognizing the novel source images; 2) By comparing the results of St-Net, Tt-Net, and M-base, we find that St-Net performs best since it is totally trained using source data. The Tt-Net performs worst and the performance of the M-base also has a degradation compared to that of St-Net. This indicates that merging the target data into the source data is harmful to the source domain. However, this negative effect is addressed by our ME-D2N. We even improve the St-Net by 2.94\% on average. This shows that decomposing the student network makes our model has sufficient capacity to learn both the knowledge of the source and target domains.

\subsection{Ablation Studies.}
\noindent\textbf{Ablation Study on Network Modules.}
The effectiveness of our main technical contributions -- multi-expert learning and domain decomposition module are studied. Results of the 5-way 5-shot setting are provided in Table~\ref{tab:modules}. We use the ``ME'' to refer to our model learned under the umbrella of multi-expert learning without applying domain decomposition to the student model. Correspondingly, the ``ME + D2N'' denotes the model equipped with both the multi-expert learning and domain decomposition module. Thus, ``ME + D2N'' also equals our full ME-D2N network. Comparing ``ME'' against the M-base, we notice that the multi-expert learning strategy improves the  M-base by up to 3.95\% on average. This illustrates that we do alleviate the data imbalance problem of the M-base. Similarly, the effectiveness of our domain decomposition module can be drawn through the advantages of ``ME + D2N'' over ``ME''.

\noindent\textbf{Ablation Study on the Number of Decomposed Blocks.} 
As stated in Sec~\ref{sec:setup}, we divide our embedding module into four blocks, thus decomposing which blocks is an important question.
To that end, we conduct experiments of decomposing different number of blocks and give the 5-way 5-shot results in Table~\ref{tab:abla}. 
As indicated in previous work~\cite{bau2017network}, the low-level filters convey more generic information thus naturally entangled, while the high-level filters are more semantic-related and much easier to be decomposed. Thus, the upper blocks are decomposed with higher priorities.
Take ``3 decomposed blocks'' as an example, it means the last three blocks are decomposed.
Results show that our choice of ``2 decomposed blocks'' performs best, then follows the ``1 decomposed block'', ``3 decomposed blocks'', and ``4 decomposed blocks'' consecutively. 
This phenomenon basically keeps the consistency of the conclusions as in ~\cite{bau2017network}. Different from those works strive for a more interpretable network that only decomposes the last semantic layer~\cite{liang2020training, zhang2018interpretable, shen2021interpretable}, in this paper, decomposing the last two blocks is the best choice for us. This reveals that the domain information is also conveyed in the relatively low-level filters.

	\begin{table*}[t]
		\centering
		\begin{tabular}{cccccc}
			\hline
			\textbf{Choices} & \textbf{CUB} & \textbf{Cars} & \textbf{Places} & \textbf{Plantae}& \textbf{Avg.} \\
			\hline
					4 decomposed blocks &  78.05 \mypm 0.62 & 63.27 \mypm 0.67 & 77.53 \mypm 0.64 & 67.85 \mypm 0.72 & 71.68 \\
			
			3 decomposed blocks & 81.76 \mypm 0.57 & 64.90 \mypm 0.67 & 78.55 \mypm 0.62 & 70.05 \mypm 0.70 & 73.82\\
			
			\textbf{2 decomposed blocks (ours)} & \textbf{83.17 \mypm 0.56}  & \textbf{69.17 \mypm 0.68} & \textbf{80.45 \mypm 0.62} & \textbf{72.87 \mypm 0.67}  &\textbf{76.42}\\
			
			1 decomposed block & 81.39 \mypm 0.60 & 68.27 \mypm 0.70 & 80.39 \mypm 0.62 & 72.18 \mypm 0.67 & 75.56\\
			\hline
			STD path & \textbf{83.32 \mypm 0.58}  &	67.85 \mypm 0.68 &	\textbf{80.60 \mypm 0.61} & 72.52 \mypm 0.70  & 76.07\\
			
			DSG path & 78.15 \mypm 0.62 & 65.37 \mypm 0.67  & 75.15 \mypm 0.67  &	68.85 \mypm 0.70 &  71.88\\
			
			\textbf{STD + DSG paths (ours)} &  83.17 \mypm 0.56 & \textbf{69.17 \mypm 0.68} & 80.45 \mypm 0.62  &	\textbf{72.87 \mypm 0.67} &  \textbf{76.42}\\
			\hline
	
		\end{tabular}
		\caption{Ablation studies of our method. We conduct experiments of decomposing filters on different numbers of blocks and using different testing strategies for model inference. Results of 5-way 5-shot tasks are reported.}
		\vspace{-0.25in}
		\label{tab:abla}
	\end{table*}

\noindent\textbf{Ablation Study on Testing Strategies.}
Recall that ME-D2N has two forward paths -- STD path and DSG path,
thus we report the results of different testing strategies in Table~\ref{tab:abla}.
The ``STD path'' and ``DSG path'' mean only a single forward path is used while the ``STD + DSG paths'' denotes both of them are utilized.
From the results, it can be observed that the STD path performs better than the DSG path. This is not difficult to understand since the STD path receives knowledge from both domains. However,the contribution of our D2N still hold as the STD path is hosted under the D2N module. 
On average, utilizing two paths generally improves the final results.

\subsection{More Analysis}
\begin{figure}[h!]
  \centering
  \vspace{-0.1in}
  \includegraphics[width=0.9\linewidth]{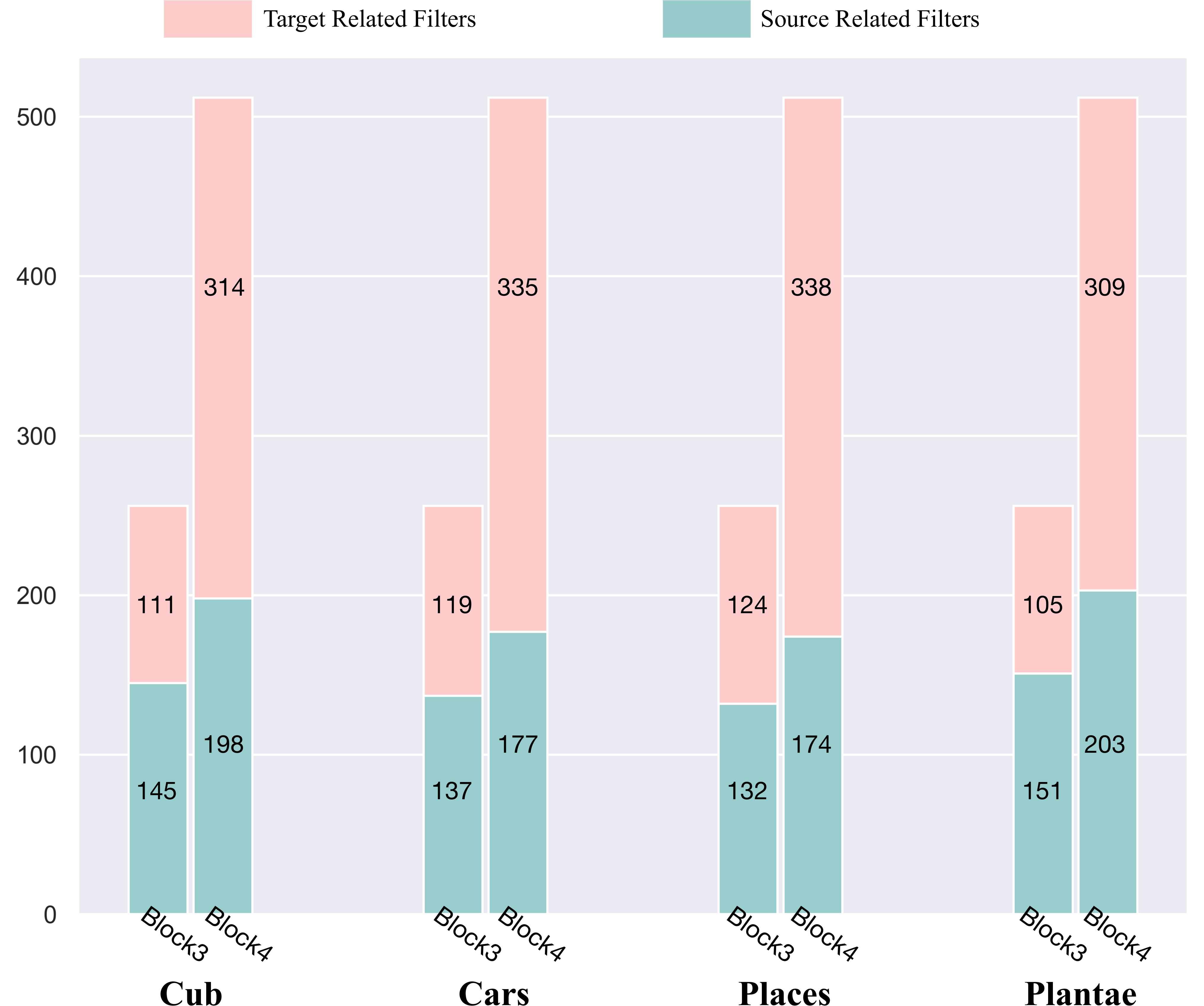}
  \vspace{-0.1in}
  \caption{The number of the filters assigned for each domain.}
  \vspace{-0.15in}
  \label{fig:filterNum}
\end{figure}

To provide more analysis of our domain decomposition module, we show the number of filters assigned for different domains in Figure~\ref{fig:filterNum}. Typically, block3 and block4 have a total of 256 and 512 filters, respectively. Firstly, we observe that the models trained on different target sets share similar distribution of the decomposed filters. Secondly, we notice that the number of target-specific filters is comparable with that of source-specific ones with slight advantages at block3. While the filters of block4 have an obvious bias towards the target domain with a rough ratio of 3:2 for target: source. This illustrates that our D2N module learns to assign more capacity to the target domain.

\begin{figure}[htb]
  \centering
   \vspace{-0.1in}
  \includegraphics[width=0.9\linewidth]{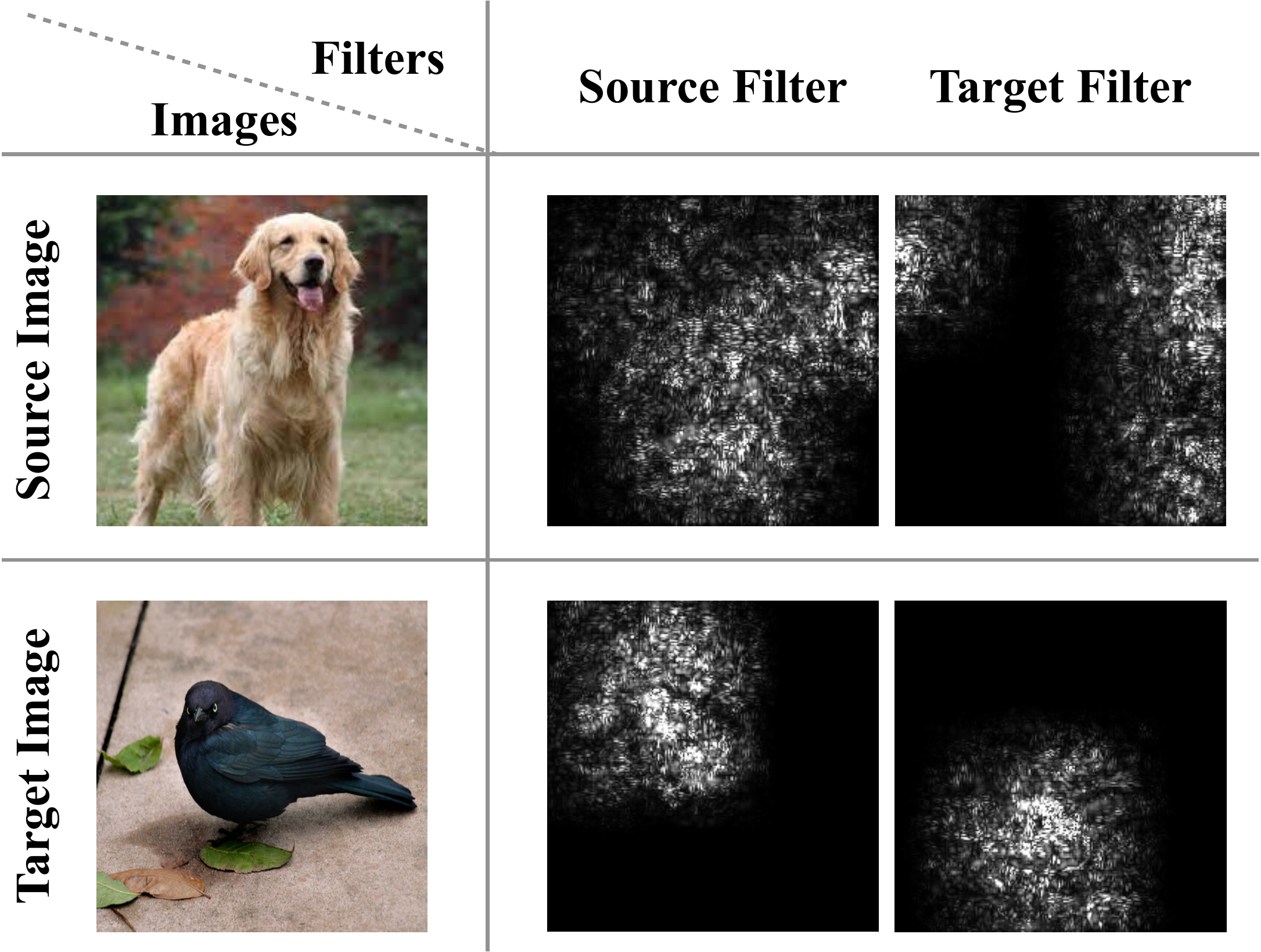}
  \vspace{-0.1in}
  \caption{
  Visualization of activation maps for input images on the source and target-related filters.
  }
  \vspace{-0.15in}
  \label{fig:filteractive}
\end{figure}

In addition, to have an intuitive understanding of our D2N,
we visualize the activation maps of two different domain-specific filters on two different domain images. 
The demonstrated ``source filter'' and ``target filter'' are sampled from the last block of the embedding module.
As shown in Figure~\ref{fig:filteractive}, the activation results of these two filters towards the same input image are significantly different.
The domain-specific filters can accurately focus on the effective features for the input image of the same domain.
This further verifies the effectiveness of our D2N module.

\section{Conclusion}
To conclude, we mainly aim at promoting the CD-FSL methods with few labeled target data. To achieve this, we first observe two key challenges lay in this task -- data imbalance issue and learning model from two different domains.
To address these problems, we thus contribute two novel modules. One is the multi-expert learning mechanism together with the knowledge distillation technique, which enables us to learn ``individual knowledge'' from two teacher models of different domains rather than learning from the ``unbalanced data''.
Another is the domain decomposition module which learns to decompose the filters of our student model into the source-specific and target-specific sub-parts. In this way, we prevent our model from learning knowledge of the source domain and target domain at the same time.
Based on these two modules, we build our multi-expert domain decompositional network. Experimental results show that our network alleviates the above-mentioned challenges well and achieves state-of-the-art results.

\section{Acknowledgement}
This work was supported in part by National Natural Science Foundation of China Project (No. 62072116) and Shanghai Pujiang Program (No. 20PJ1401900).

\bibliographystyle{ACM-Reference-Format}
\balance
\bibliography{references}

\end{document}